\PassOptionsToPackage{colorlinks=false,allbordercolors={1 1 1}}{hyperref}
\documentclass[twoside,11pt]{article}

\usepackage[utf8]{inputenc}
\usepackage{blindtext}

\usepackage{amsmath}   % Essential for most math environments
\usepackage{amssymb}   % Provides extra mathematical symbols as \mathbb{E}
\usepackage{mathtools} % Extends amsmath and fixes minor bugs
\usepackage{amsfonts}  % Required for math fonts
\usepackage{mathrsfs}  % Optional: if you prefer a different script H
\usepackage{bm}        % For bold math symbols
\usepackage{algorithm}
\usepackage{algpseudocode} % This provides the algorithmic environment
\usepackage{float}

\usepackage{tabularx}
\usepackage{booktabs} % For professional looking lines
\usepackage[table]{xcolor} % If you want to add the grey shading
\usepackage{epsfig}
\usepackage{epstopdf}
\definecolor{lightgray}{gray}{0.95}

\usepackage{graphicx}
\graphicspath{ {./Figures/} }
\usepackage{subcaption}

\usepackage{jmlr2e}

\usepackage{lastpage}
\ShortHeadings{Hierarchical Soft Actor-Critic for Sparse-Reward Long-Horizon Reinforcement Learning}{Elashaal, Hfaiedh, Khraief, Ellabib and Cirrincione}
\firstpageno{1}

\begin{document}

\title{Hierarchical Soft Actor-Critic for Sparse-Reward Long-Horizon Reinforcement Learning}

\author{\name Zahra Abdalla Elashaal    \email z.elashaal@uot.edu.ly \\
       \addr Lab. of Robotics, Informatics and Complex Systems, \\National Engineering School of Tunis, University of Tunis El Manar,\\ Tunis, Tunisia\\
       \addr Faculty of Information Technology, University of Tripoli, \\ Tripoli, Libya \\
       \name Afef Hfaiedh \email afef.hfaiedh@enit.utm.tn \\
       \addr Lab. of Robotics, Informatics and Complex Systems, \\National Engineering School of Tunis, University of Tunis El Manar,\\ Tunis, Tunisia\\       
       \name Nahla Khraief \email nahla.khraief@enit.utm.tn \\
       \addr Lab. of Robotics, Informatics and Complex Systems, \\National Engineering School of Tunis, University of Tunis El Manar,\\ Tunis, Tunisia\\
       \name Issmail Ellabib  \email i.ellabib@uot.edu.ly \\
       \addr Computer Engineering Dept.,\\ University of Tripoli\\ Tripoli, Libya \\
       \name Giansalvo Cirrincione \email giansalvo.cirrincione@u-picardie.fr \\
       \addr Laboratory of Novel Technologies,\\      University of Picardie Jules Verne\\
       Amiens, France
}

\editor{}
\maketitle 

\begin{abstract}%   <- trailing '%' for backward compatibility of .sty file
Exploration in sparse-reward long-horizon tasks poses significant challenges for reinforcement learning. To address these challenges, we propose a two-level Hierarchical Reinforcement Learning (HRL) framework. The first level handles high-level strategic planning, while the low-level uses the continuous-control Soft Actor-Critic (SAC) algorithm, and they utilize entropy-regularized policy optimization. The proposed framework was trained and evaluated using the Search-and-Rescue-2 (SAR-2) dataset. HRL-SAC effectively addresses sparse-reward long-horizon search problems characterized by delayed rewards and continuous control, and its outperforming the flat SAC baseline reinforcement learning in terms of success rates, coverage efficiency, and convergence. These findings indicate that hierarchical entropy-regularized policies are a promising solution to tackle long-horizon sparse-reward reinforcement learning tasks.
%\blindtext
\end{abstract}



\section{Introduction}\vspace{-0.35cm}
Reinforcement Learning (RL) has emerged as a powerful paradigm for training autonomous agents to solve complex sequential decision-making tasks through environmental interaction \cite{sutton2018reinforcement}. By maximizing long-term cumulative returns, RL agents learn to map environmental observations to optimal actions across diverse domains ranging from game playing to robotics. However, standard deep RL algorithms often struggle in environments characterized by sparse extrinsic rewards and long horizons, where the probability of randomly encountering rewarding states is exponentially low. In such settings, conventional exploration strategies frequently fail to propagate credit back to early decision points, leading to stalled convergence or sub-optimal policies.

Hierarchical Reinforcement Learning (HRL) addresses these limitations by introducing temporal and behavioral abstraction. By decomposing a complex task into a hierarchy of sub-problems, HRL enables agents to reason over extended time scales. Typically, a high-level policy (meta-controller) selects long-term sub-goals, while a low-level policy (worker) executes primitive actions to satisfy these objectives. Despite the theoretical appeal of HRL, combining hierarchical structures with continuous control and off-policy learning remains challenging due to non-stationary transition distributions and unstable gradients.

\subsection{Related Works}
This section will explore the most relevant studies on Hierarchical Reinforcement Learning (HRL) that focus on Long-Horizon Path Planning and Entropy-Regularized Reinforcement Learning.

\noindent
{\bf Hierarchical reinforcement learning} has long been studied as a means of addressing the challenges of long-horizon decision making and sparse rewards by introducing temporal abstraction. Early frameworks, such as the option framework, formalized temporally extended actions and established a theoretical foundation for hierarchical policies \cite{Sutton1999}. Subsequent work explored hierarchical value functions and semi-Markov decision processes to improve scalability in complex environments \cite{Dietterich2000}. 
More recent methods utilize deep neural networks to learn hierarchical policies in end-to-end manner, such as Feudal RL, \cite{Vezhnevets2017}, and option-critical architectures \cite{Bacon2017} enable joint learning of high-level decision making and low-level control. Although these methods demonstrate improved exploration, they often rely on discrete action spaces or lack stability in continuous control settings. Hierarchical methods have also been applied to sparse-reward domains, where high-level policies guide exploration, while low-level controllers handle execution \cite{Kulkarni2016}. However, many existing HRL approaches remain sensitive to reward design or struggle with unstable training dynamics, particularly in continuous action spaces. Recent advances in HRL have demonstrated its effectiveness in addressing sparse rewards and long-horizon decision-making in complex environments. Modern HRL approaches build upon earlier temporal abstraction frameworks while leveraging deep neural networks for scalable policy learning. Methods such as the Hierarchical Actor-Critic (HAC) \cite{Levy2017} and Hierarchical Reinforcement learning
with Off-policy correction (HIRO-A) algorithm \cite{Chang2025}  decompose decision-making into high-level goal selection and low-level continuous control, allowing improved exploration and sample efficiency. Other recent work as \cite{yan2024hierarchical} addresses the problem of navigation with sparse rewards with Hierarchical RL with Multi-Goal (HRL-MG) but this hierarchical architecture is designed to specify the end goal based on the discrete action space. 
Hierarchical reinforcement learning has been extensively studied as a framework for introducing temporal abstraction into sequential decision making. Comprehensive surveys summarize a wide range of hierarchical formulations, including options, goal-conditioned policies, and multi-level controllers, as well as their theoretical and empirical properties \cite{al2015hierarchical,pateria2021hierarchical,hutsebaut2022hierarchical}. These surveys identify long-horizon credit assignment and exploration under sparse rewards as persistent open challenges, particularly in continuous control and off-policy learning settings.
Another study, while not HRL, divides the learning challenge into action space learning and policy learning in the context of RL, similar to the LASER method \cite{allshire2021laser}. A different model focuses on control systems defined in infinite-dimensional function spaces, proposing RL architecture algorithms for optimal policy learning across large agent systems \cite{zhang2025reinforcement}. Unlike standard RL agents that map states to actions, HRL agents create high-level policies for sub-goals and low-level policies for execution, enhancing exploration and credit assignment.

\noindent
{\bf Hierarchical Reinforcement Learning with Continuous Control,} Recent works have explored combining hierarchical structures with continuous control policies. Methods such as Hierarchical Actor-Critic \cite{Nachum2018} demonstrate that decomposing tasks into high-level goals and low-level controllers can significantly improve learning efficiency in long-horizon tasks. These methods show promising results, but often rely on deterministic low-level policies or handcrafted goal representations. Moreover studies have explored HRL in continuous and high-dimensional domains. \cite{liu2021hierarchical} demonstrate that hierarchical structures hierarchical deep reinforcement learning with automatic sub-goal identification (HADS) improved performance in sparse-reward environments. Despite these advances, training instability and inefficient exploration remain challenges, particularly when scaling HRL to long-horizon continuous control problems. Other works explored incorporate hierarchical structures with continuous control and off-policy learning, and intrinsic rewards to guide low-level and high-level policies operate. \cite{Levy2017} introduce HAC methods that enable end-to-end training of hierarchical policies in continuous domains. While \cite{pmlr-shen20b}, demonstrate the smooth regularized reinforcement learning effectively constrains the search space and enforces smoothness in the learned policy.  However, in its effectiveness, this method may suffer from instability when scaling to complex environments or when reward signals are highly sparse. 
In addition, the study \cite{xu2023} explores hierarchical reinforcement learning with entropy-regularized objectives, showing promising results in long-horizon tasks. Nevertheless, these approaches often rely on deterministic low-level policies or require complex architectural tuning.
More recently, some prior work has explored HRL for long-horizon path planning in complex robotic systems. Distributed hierarchical frameworks have been proposed for micro-assembly tasks, demonstrating improved scalability through task decomposition and coordination between multiple controllers \cite{Zhao2025}. \cite{LV2025} presents hierarchical Twin-delayed deep deterministic (H-TD3) algorithm, a long-horizon path planning with a hierarchical framework operating on different temporal scales. Similarly, hierarchical approaches have been applied to stratospheric airship navigation to address extended planning horizons and delayed rewards \cite{yang2025actor}. While these methods highlight the effectiveness of hierarchical decomposition in long-horizon settings, they typically rely on task-specific designs or alternative optimization objectives, and do not explicitly incorporate entropy-regularized off-policy learning. 
Recent work such as \cite{zeng2025hierarchical} proposes a hierarchical framework to improve decision-making efficiency in complex environments. The authors design a structured multi-level architecture that decomposes high-level objectives into low-level executable actions, improving scalability and stability.
As HRL agents merely optimize for rewards, the agent tends to search the same space redundantly, which reduces the speed of learning. \cite{watanabe2022shiro} present an Off-Policy Soft Hierarchical Reinforcement Learning (SHIRO) algorithm that maximizes entropy for efficient exploration. 
Compared to their approach, our proposed framework integrates entropy-regularized Soft Actor-Critic optimization at both meta and low-level controllers, enabling improved exploration and sample efficiency in autonomous search and rescue tasks.

\noindent
{\bf Actor-Critic and Entropy-Regularized Reinforcement Learning,} SAC is an off-policy actor-critic algorithm that optimizes a maximum-entropy objective, encouraging stochastic policies that balance exploration and exploitation. By incorporating an entropy term into the reward function, SAC achieves improved stability and sample efficiency compared to deterministic policy gradient methods and become a widely adopted algorithm for continuous control due to its robustness and sample efficiency. SAC has demonstrated strong performance on continuous control benchmarks and robotic manipulation tasks \cite{Haarnoja2018applications}. However, flat SAC policies often struggle in sparse-reward and long-horizon environments due to limited temporal abstraction and inefficient exploration. 
Entropy-regularized RL has been shown to significantly improve exploration and stability in continuous control tasks. Although entropy regularization promotes local exploration, it does not fully address the challenge of delayed credit assignment over extended time horizons. Several extensions have attempted to improve SAC’s exploration capabilities through intrinsic motivation or auxiliary objectives \cite{Burda2018}. Recent extensions of SAC focus on improving exploration in sparse-reward settings through intrinsic motivation, representation learning, or auxiliary objectives \cite{Laskin2020}. However, these approaches typically operate at a single temporal scale and do not explicitly model hierarchical structure and long-horizon credit assignment. SAC formulates policy optimization using a maximum-entropy objective and enables efficient off-policy learning using stochastic policies \cite{hou2020off}. Although SAC demonstrates strong performance on a wide range of benchmarks, it operates as a flat policy and does not explicitly address long-horizon decision making or sparse-reward credit assignment through temporal behaviors. A similar challenge in autonomous navigation is addressed by \cite{abdallah2025aidriven}, they introduced a Deep RL framework for real-time Unmanned Aerial Vehicle (UAV) trajectory optimization in dynamic and partially observable environments. Their work utilizes the Advantage Actor-Critic (A2C) algorithm to navigate 3D spaces, focusing on robust obstacle avoidance and computational efficiency. Their approach demonstrates high adaptability in stochastic settings, focuses on aerial platforms using A2C, while our work extends these principles by using a Hierarchical Soft Actor-Critic  architecture, which handles the sparse-reward and long-horizon nature of Search-and-Rescue missions.  Also, to balance between exploration and exploitation, a historical decision-making regularized maximum entropy (HDMRME) RL algorithm is developed by \cite{dong2024historical}, but this algorithm did not applied in autonomous driving. 
Another important categorization is based on the learning objective. Value-based methods, such as Deep Q-Networks (DQN) , aim to estimate value functions (e.g., action-value functions) and derive the policy implicitly by selecting actions that maximize the estimated value \cite{gio2025}. Policy-based methods, such as REINFORCE (Policy Gradient), directly optimize the policy without explicitly learning a value function. Also, actor–critic approaches in \cite{actorcriticjmlr2025}, including Deep Deterministic Policy Gradient (DDPG) \cite{afef2024} and Proximal Policy Optimization (PPO) \cite{schulman2017ppo}, combine both paradigms by using an actor network to represent the policy and a critic network to estimate the value function, thereby improving learning stability and efficiency.

The effective learning of policies in long-horizon, sparse-reward reinforcement learning environments remains a central open problem in machine learning. Standard deep reinforcement learning methods often struggle to propagate delayed rewards over extended temporal horizons, leading to inefficient exploration and unstable convergence. Hierarchical reinforcement learning addresses this challenge by introducing temporal abstraction, enabling agents to reason over multiple time scales.

\textbf{Our contribution:}  
Unlike prior hierarchical RL approaches that either rely on deterministic low-level policies as HAC by \cite{Nachum2018}, discrete action spaces such as HRL-MG by \cite{yan2024hierarchical}), or task-specific heuristics, our Hierarchical RL of Soft Actor-Critic (HRL-SAC) framework introduces a novel integration of entropy-regularized Soft Actor-Critic optimization across both hierarchical levels within a continuous control setting. While existing SAC extensions address exploration in flat policies, they hard to handle long-horizon sparse rewards through temporal abstraction; similarly, other HRL methods like SHIRO \cite{watanabe2022shiro} apply soft hierarchies but lack the dual-level entropy regularization and goal-conditioned SAC that enable our approach's superior sample efficiency and convergence. This dual-entropy HRL formulation provides the theoretically-grounded solution combining SAC's maximum-entropy stability with hierarchical credit assignment, without auxiliary rewards or architectural tuning. The proposed approach applies entropy regularization at multiple temporal levels, enabling stable off-policy learning and exploration robustness in sparse-reward long-horizon environments. This formulation preserves the theoretical benefits of SAC while addressing delayed credit assignment through hierarchical decomposition, without relying on auxiliary rewards or task-specific heuristics.

\begin{figure}[h]
    \centering
    \includegraphics[width=.6\textwidth]{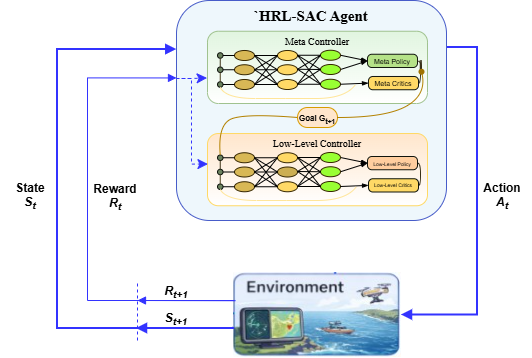}
    \caption{HRL-SAC Agent Architecture. The meta-controller generates high-level sub-goals based on environment observations, while the low-level controller executes entropy-regularized actions conditioned on these sub-goals.}
    \label{fig:HRL Agent}
\end{figure}

The proposed HRL-SAC agent is illustrated in Figure~\ref{fig:HRL Agent}. The overall architecture decomposes decision making into meta (high-level) and low-level control layers interacting with a shared environment. The meta-controller operates at a coarser temporal scale and receives the global state from the environment. It consists of a meta-policy and meta-critic networks, which jointly learn to produce a high-level sub-goal $g_t$. This sub-goal represents an abstract objective or behavioral directive that guides the lower layer over multiple time steps. The low-level controller operates at a finer temporal resolution and is conditioned on both the current state and the sub-goal $g_t$  generated by the meta-controller. It comprises a low-level policy and low-level critic networks, which produce primitive continuous actions executed directly in the environment. The low-level controller is optimized using the Soft Actor–Critic framework to maximize entropy-regularized returns while satisfying the assigned sub-goal.
The environment provides observations and rewards to both hierarchical levels. Low-level actions influence the environment dynamics, while accumulated task feedback is used to update both the low-level SAC components and the high-level meta-policy. This hierarchical interaction enables efficient learning in long-horizon, sparse-reward tasks by separating strategic planning from reactive control.

\section{Theoretical Foundations of Hierarchical Soft Actor-Critic}

Reinforcement learning problems are commonly formulated as Markov Decision Processes (MDPs) \cite{sutton2018reinforcement}. While modern deep RL methods have achieved impressive results in continuous control, flat policy optimization remains sample-inefficient in sparse-reward and long-horizon environments due to delayed credit assignment. We consider MDP defined by the tuple: 
   \[M=(S,A,p,r,\gamma], \]

Where $S$ denotes the state space, $A$ the continuous action space, $p$ is the transition kernel,  \(r\) is the reward function, and \(\gamma \in (0,1)\) the discount factor. In sparse-reward long-horizon environments, direct optimization of flat policies is often inefficient due to delayed credit assignment.
HRL addresses this limitation by introducing temporal abstraction, enabling high-level decision making over extended time scales. Separately, entropy-regularized methods such as Soft SAC provide stable and efficient learning by explicitly encouraging stochastic exploration. 
 \begin{equation}
 \pi^* = \arg \max_{\pi} \mathbb{E}_{\tau \sim \pi} \left[ \sum_{t=0}^{T} r(s_t, a_t) + \alpha \mathcal{H}(\pi(\cdot \mid s_t)) \right] 
 \label{eq:SAC_policy}
\end{equation}
where \(\pi^*\) is an optimal policy that maximizes the return, \(\mathbb{E}_{\tau \sim \pi}\) is expected value or average future value after the transition, \(r(s_t, a_t)\) is a dynamic reward depending on the state $s_t$ and the action $a_t$ at time $t$, \(\mathcal{H}\) is the entropy term, \(\alpha\) is the temperature parameter controlling the trade off between maximizing the expected return and promoting exploration through higher policy entropy.
SAC optimizes a maximum-entropy objective that increases the expected return with an entropy regularization term, encouraging exploration and robustness \cite{hutsebaut2022hierarchical}.
The \(Q(s, a)\) denotes the expected long-term return of the current state is expressed as:

\begin{equation}
Q(s, a) = r(s, a) + \gamma \mathbb{E}_{s'} [V(s')] 
\end{equation}

The value function \(V(s) \) is defined as follows:
\begin{equation}
V(s) = \mathbb{E}_{a \sim \pi} [Q(s, a) - \alpha \log \pi(a \mid s)]
\end{equation}
Soft Bellman backup incorporates policy entropy directly into the value estimation, resulting in improved stability in continuous control settings \cite{Haarnoja2018applications}. By combining these two principles into HRL-SAC framework that leverages temporal abstraction and entropy maximization to improve learning efficiency in sparse-reward long-horizon tasks.
To address this challenge, in our proposed HRL-SAC framework we decomposed decision-making into two temporal levels: a high-level meta-policy responsible for long-horizon planning, and a low-level Soft Actor-Critic policy responsible for continuous control.

\subsection{Hierarchical Policy Structure}
The hierarchical policy consists of:
\begin{itemize}
    \item A (Meta-Agent) uses a stochastic meta-policy \(\pi_\phi(g|s) \) that generates a sub-goal $g\in G$   every $K$ time step, $\phi$ the weights of the high-level policy network, this process handled by the Meta Actors (which likely outputs parameters for a Gaussian distribution). 
    \item a low-level policy \(\pi_\theta(a|s,g) \) that produces continuous actions conditioned on the current state and the generated sub-goal $g$.
\end{itemize}
At time step t, the execution proceeds as:
\begin{equation}
g_t \sim \pi_\phi(g|s_t) \qquad  a_{t+i}\sim  \pi_\theta(a|s_{t+i},g_t); \qquad i=0,...,K-1. 
\end{equation}
This temporal abstraction allows the meta-policy to reason over extended horizons, while the low-level policy focuses on short-term control.

\subsubsection{Low-Level Soft Actor-Critic Objective}
The low-level controller is trained using Soft Actor-Critic, which optimizes a maximum entropy objective. The objective for the low-level policy is defined as:
\begin{equation}
    J_{\text{LL}}(\pi_\theta) = \mathbb{E}_{(s,a,g) \sim \mathcal{B}} \left[ Q_\psi(s, a, g) - \alpha \log \pi_\theta(a \mid s, g) \right],
\end{equation}
where; \(J_{\text{LL}}\) is the low-level Loss function, $Q_\psi$ denotes the soft Q-function, $\alpha > 0$ is the temperature parameter controlling the entropy-regularization trade-off, and $\mathcal{B}$ is the replay buffer.
The corresponding Bellman backup for the soft Q-function is:
\begin{equation}
    Q_\psi(s, a, g) = r_{\text{int}}(s, a, g) + \gamma \mathbb{E}_{s' \sim p} \left[ V_{\bar{\psi}}(s', g) \right],    
\end{equation}
; \(r_{\text{int}}\) intrinsic reward
with the soft value function defined as:
\begin{equation}
    V_{\bar{\psi}}(s, g) = \mathbb{E}_{a' \sim \pi_\theta} \left[ Q_{\bar{\psi}}(s', a', g) - \alpha \log \pi_\theta(a' \mid s', g) \right].
\end{equation}

\subsubsection{Intrinsic Reward for Sub-Goal Execution}
To guide the low-level policy toward accomplishing sub-goals selected by the meta-policy, an intrinsic reward is defined as:
\begin{equation}
   r_{\text{int}}(s_t, a_t, g_t) = -\| f(s_{t+1}) - g_t \|_2, 
   \label{eq:int_reward}
\end{equation}
where;   \(||.||_2\) is the $L_2$ norm, $f(\cdot)$ is a mapping function (like a coordinate extractor) that aligns the environment state with the sub-goal space to get goal-relevant representation.
This intrinsic reward encourages the low-level controller to minimize the distance between the achieved state and the desired sub-goal, facilitating efficient sub-goal execution and stable learning.

\subsubsection{High-Level Meta-Policy Optimization}
The high-level meta-policy operates over a temporally extended MDP, where each action corresponds to selecting a sub-goal $g$, and transitions occur every $K$ steps. The transition dynamics \( p(s'|s,a) \) for the High-Level agent  is not standard MDP transitions because they are temporally extended. The transition is defined as \( (s_t,z_t,R_t,s_{t+K})\), where $R_t$ is the cumulative environment reward or extrinsic return for a meta-action and defined as:
\begin{equation}
   R_t^{\text{meta}} = \sum_{i=0}^{K-1} \gamma^i r_{\text{ext}}(s_{t+i}, a_{t+i}), 
   \label{eq:Meta_Reward}
\end{equation}
where $r_{\text{ext}}$ is the reward for the sparse task, $\gamma$ is the discount factor between 0-1.
The meta-policy is optimized to maximize expected extrinsic return:
\begin{equation}
   J_{\text{HL}}(\pi_\phi) = \mathbb{E}_{(s,g) \sim \mathcal{B}_{\text{meta}}} [Q_\omega^{\text{meta}}(s, g)], 
\end{equation}

with the meta-level Bellman update:
\begin{equation}
   Q_\omega^{\text{meta}}(s, g) = R_t^{\text{meta}} + \gamma^K \mathbb{E}_{s' \sim p} \left[ \mathbb{E}_{g' \sim \pi_\phi} Q_\omega^{\text{meta}}(s', g') \right]. 
\end{equation}

where; \(Q_\omega^{\text{meta}}(s, g)\) the meta $Q$ value estimates expected cumulative return of selecting sub-goal $g$ in state $s$, $\omega$ is the weights of the meta-level critic network, \(R_t^{\text{meta}}\) Accumulated extrinsic reward collected by the low-level policy over the execution of sub-goal $g$.

The effectiveness of HRL-SAC in sparse-reward long-horizon environments arises from the combination of: Temporal Abstraction for shorter effective planning horizons, Intrinsic rewards give dense learning signals, Entropy regularization which stabilizes policy optimization and improves exploration,  Goal-conditioned control, which enables reusable low-level behaviors. and Hierarchical decomposition reduces gradient variance. This is a claim\footnote{Supported by empirical results in Section \ref{sec:Results} and HRL surveys [ \cite{pateria2021hierarchical}; \cite{hutsebaut2022hierarchical}]}.
 Together, these properties show why HRL-SAC consistently outperforms flat RL methods in challenging sparse-reward settings.

\subsection{Sparse Rewards and Credit Assignment}
Learning in sparse-reward, long-horizon RL environments is fundamentally challenging due to delayed credit assignment, inefficient exploration, and high variance in policy gradient estimates. This section provides theoretical intuition and analysis that explains how the proposed HRL-SAC framework alleviates these challenges.

In flat reinforcement learning, the expected return of a policy \(\pi\) is given by:
\begin{equation}
 J(\pi) = \mathbb{E}_{\tau \sim \pi} \left[ \sum_{t=0}^{T} \gamma^t r(s_t, a_t) \right]. 
\end{equation}
When rewards are sparse, non-zero signals occur only after long trajectories, making gradient estimates:
\[ \nabla_\theta J(\pi_\theta) \]
high-variance and difficult to optimize. As the effective horizon $T$ increases, the probability of randomly discovering rewarding trajectories decreases exponentially. Hierarchical decomposition reduces this difficulty by introducing intermediate decision points that receive denser learning signals, shortening the effective planning horizon.

Temporal abstraction reduces effective horizon, by selecting sub-goals every $K$ time step, HRL-SAC induces a semi-Markov decision process (SMDP) at the meta-level. The meta-policy optimizes:
\begin{equation}
    J_{\text{meta}} = \mathbb{E} \left[ \sum_{t=0}^{T/K} \gamma^{tK} R_t^{\text{meta}} \right],
\end{equation}

Where $R_t^{meta}$ aggregates the rewards in $K$ steps. This intuitively reduces the effective horizon from $T$ to $T/K$  at the meta-level, providing theoretical motivation for improved sample efficiency and credit assignment across long trajectories, as observed empirically, and reduces gradient variance. So that, the meta-policy learns strategic behaviors more efficiently than flat policies.

Intrinsic rewards densify the learning signal, and the low-level policy receives an intrinsic reward using the equation 
\eqref{eq:int_reward}
which provides a dense learning signal independent of task-level rewards. This intrinsic reward ensures continuous feedback even when extrinsic rewards are absent, stabilizes early-stage learning, and enables low-level policy to learn reusable skills. 
From a learning theory perspective, intrinsic rewards can reduce the variance of value estimation by replacing sparse, delayed rewards with dense signals, consistent with the shaped reward analyses in RL theory \cite{ng1999policy}.

\subsection{Entropy Regularization Improves Exploration}
Soft Actor-Critic optimizes a maximum-entropy objective $J_{\text{SAC}}$ that is shown in equation \eqref{eq:SAC_policy},
which encourages stochastic policies and prevents premature convergence. In HRL-SAC, entropy regularization improves robustness to sub-optimal meta-policy decisions. promotes exploration within sub-goal execution and avoids deterministic collapse of low-level behaviors. This entropy-driven exploration complements hierarchical structure by ensuring diversity at the execution level.
A key advantage of HRL-SAC is the \textbf{separation of concerns} or the decoupling of exploration and exploitation across levels.
The meta-policy focuses on long-term exploration and strategy, and the low-level policy focuses on short-term control and stability.
This decoupling allows each policy to operate at an appropriate temporal scale, reducing interference between long-horizon planning and short-horizon control. The resulting optimization landscape is smoother and easier to optimize compared to flat policies.

Reduction in the gradient variance of the policy. Let $\hat{g}_{\text{flat}}$ and $\hat{g}_{\text{HRL}}$ denote gradient estimators for flat and hierarchical policies, respectively. Due to reduced horizon and denser rewards, HRL-SAC intuitively yields the following requirements:

\begin{equation}
    \text{Var}(\hat{g}_{\text{HRL}}) < \text{Var}(\hat{g}_{\text{flat}}),
\end{equation}
${Var}$ is the variance that indicates how much the returns fluctuate between episodes. As HRL has a lower variance, leading to more stable learning and faster convergence. Although formal variance bounds are difficult to derive in general settings, empirical evidence in the HRL literature \cite{pateria2021hierarchical} and our experiments supports this intuition.

\section{Experimental Setup}
The proposed method is evaluated in a simulated search task that exhibits sparse rewards, partial observability, and long-horizon decision-making. Although the environment is inspired by search-and-rescue scenarios, it primarily serves as a benchmark for evaluating hierarchical reinforcement learning under challenging conditions.

\subsection{Environment}
In order to evaluate the proposed HRL-SAC framework, experiments are conducted in a simulated continuous-control environment designed to exhibit key challenges of sparse-reward long-horizon reinforcement learning. The environment exhibits delayed extrinsic rewards, stochastic transitions, and partial observability, which require agents to explore efficiently across extended time horizons. Although inspired by search tasks, the environment primarily serves as a benchmark for evaluating proposed algorithm under sparse-reward conditions. domain-specific heuristics are not established in the learning process.
The state space $S$ consists of continuous-valued observations capturing the agent’s position, velocity, and task-relevant environmental features. Observations are normalized to zero mean and unit variance. The action space $A$ is continuous and bounded, representing low-level control commands. All policies output actions via squashed Gaussian distributions to ensure bounded actions.
The hierarchical configuration policy operates at two temporal resolutions: a) Meta-policy interval; here the high-level policy selects a sub-goal every $K$ environment steps, and b) Low-level control; where the SAC controller executes continuous actions conditioned on the selected sub-goal. The sub-goal space $G$ is defined in the same representation space as a subset of the state variables, allowing goal-conditioned control without handcrafted abstractions.
To assess the effectiveness of HRL-SAC, the baseline algorithm flat SAC without hierarchical decomposition is evaluated and implemented using standard, publicly available implementations and tuned to achieve the best reported performance under the same comparable training parameters.

\subsection{Datasets }
Two types of datasets are introduced to evaluate the proposed framework as follows:
\subsubsection{Synthetic Dataset}
Preliminary development with synthetic dataset: Initial development and algorithm validation of the hierarchical reinforcement learning framework was conducted using a synthetic maritime search-and-rescue dataset comprising 200 procedurally generated scenarios. This dataset was automatically created as a fallback when the primary annotation file was unavailable, featuring six target categories (person, boat, life-jacket, debris, floating object, swimmer) with randomized  placements of the bounding box on 640×480 images. Synthetic data enabled rapid prototyping, debugging, and baseline performance assessment of both HRL-SAC and standard SAC agents, revealing key insights into hierarchical coordination and providing a controlled environment for isolating algorithmic behavior prior to real-data evaluation.

\subsubsection{SAR-2 Dataset}
Real-World evaluation with SAR-2 Dataset: Comprehensive evaluation was performed using the SAR-2 (Search-and-Rescue-2) dataset, a production-grade computer vision benchmark containing 1980 images sourced from Kaggle,  \cite{gegenava2024sard2}, and originally curated via Roboflow, \cite{roboflow_sard2025}. This dataset images provides a realistic train/validation/test split of 70$\%$ / 20$\%$ / 10$\%$ respectively, with YOLO-v5-format annotations across categories if person or other objects. The agents were trained on the 1385-image training partition, with validation 396 images used for hyperparameter tuning and the held-out test set 199 images reserved for final performance reporting. This rigorous split isolation ensures unbiased evaluation of the HRL-SAC framework's transferability to real-world search-and-rescue drone operations.

\subsection{Training Procedure}
Training proceeds by alternating updates between the two levels:
\begin{enumerate}
    \item The meta-policy selects a sub-goal every $K$ steps.
    \item The low-level SAC policy executes actions conditioned on the sub-goal.
    \item Intrinsic and extrinsic rewards are stored in separate replay buffers.
    \item The low-level SAC policy is updated using intrinsic rewards.
    \item The meta-policy is updated using extrinsic returns aggregated over $K$ steps.
\end{enumerate}
This decoupled optimization stabilizes learning and enables effective long-horizon planning.
The HRL-SAC  algorithm present in Alg.~\ref{alg:hrl_sac}, combines temporal abstraction with maximum-entropy off-policy learning by jointly optimizing a high-level policy over latent sub-goals and a low-level policy over continuous actions.

\begin{algorithm}[h]
\footnotesize
\caption{Hierarchical Soft Actor-Critic (HRL-SAC)}
\label{alg:hrl_sac} 
\begin{algorithmic}[1]
\State Initialize high-level policy $\pi_H(g|s;\theta_H)$
\State Initialize low-level policy $\pi_L(a|s,g;\theta_L)$
\State Initialize Q-networks $Q_H, Q_L$ with parameters $\phi_H, \phi_L$
\State Initialize target networks $\phi_H' \leftarrow \phi_H$, $\phi_L' \leftarrow \phi_L$
\State Initialize replay buffer $\mathcal{B}$

\For{each episode}
    \State Observe initial state $s_0$
    \For{$t = 0$ to $T$}

        \If{$t \bmod K = 0$}
            \State Sample sub-goal $g_t \sim \pi_H(g|s_t)$
        \EndIf
        \State Sample action $a_t \sim \pi_L(a|s_t, g_t)$
        \State Execute $a_t$
        \State Observe reward $r_t$ and next state $s_{t+1}$
        \State Store transition $(s_t, g_t, a_t, r_t, s_{t+1})$ in $\mathcal{B}$

        \State Sample mini-batch from replay buffer $\mathcal{B}$

        \State // Low-Level Critic Update
        \State Compute target value $y_t^L = r_t + \gamma V_L(s_{t+1})$
        \State Update $\phi_L$ by minimizing soft Bellman error

        \State // Low-Level Actor Update
        \State Update $\theta_L$ using entropy-regularized objective

        \If{$t \bmod K = 0$}
            \State // High-Level Critic Update
            \State Compute high-level return $R_t = \sum_{i=t}^{t+K} r_i$
            \State Update $\phi_H$ using soft Bellman backup

            \State // High-Level Actor Update
            \State Update $\theta_H$ using entropy-regularized objective
        \EndIf

        \State // Target Network Update
        \State $\phi' \leftarrow \tau \phi + (1-\tau)\phi'$

        \State $s_t \leftarrow s_{t+1}$
    \EndFor
    \If{episode mod 50 == 0}
        \State Save $\pi_H$ weights to \texttt{./trained\_models/meta\_agent.pt}
        \State \textbf{Log:} Save Success Rate and Reward to \texttt{training\_stdout.log}
    \EndIf
\EndFor
\end{algorithmic}
\end{algorithm}

The agent's interaction with the environment is defined by a multi-dimensional observation and action space, as detailed in Table   \ref{tab:Action_env_parameters} and Table \ref{tab:Obs_env_parameters}. These  vectors are action space continuous bounded with 4 dimensions, and observation with total 276 dimensions, which incorporating both continuous physical parameters—such as wind speed and visibility—and discrete categorical data, such as weather conditions and high-level goals.
Observations form a 276-dimensional continuous vector, normalized to zero mean/unit variance. Components include agent position and target position tow diminution (2D) each, while, time progress, sea condition, weather, and time of day are 1D each, also, visibility (1D continuous 0.3-1.0), wind speed (1D 0.0-1.0), object detection binaries 6D, search grid coverage (64D from 32x32 down-sampled), and high-level goal one-hot 4D.

The full state $s \epsilon S$ matches the 276D observation vector, capturing agent pose, environmental dynamics (such as wind/visibility stochastics), detection grid, and goal encoding for both high and low-level policies. Sub-goal space $G$ subsets state variables as coordinates for L2-distance intrinsic rewards.

\begin{table}[h]
\centering \footnotesize
\caption{Action Space Environment Parameters}
\label{tab:Action_env_parameters}
\rowcolors{2}{white}{lightgray}
\begin{tabularx}{\textwidth}{l l l X}
\toprule
\textbf{Component} & \textbf{Dimension}& \textbf{Range} & \textbf{Description} \\
\midrule
Movement X	&1	&[-1, 1]	&Drone horizontal movement\\
Movement Y	&1	&[-1, 1]	&Drone vertical movement\\
Search Intensity	&1	&[0, 1]	&Sensor search effort (0=off, 1=max)\\
Camera Zoom	&1	&[0, 1]	&Optical zoom level\\
\midrule
\rowcolor{white} 
\textbf{Total}      & \textbf{4} & \textbf{Continuous bounded} & \textbf{Normalized output from tanh} \\
\bottomrule
\end{tabularx}
\end{table}

\begin{table}[h]
\centering \footnotesize
\caption{Observation Space Environment Parameters}
\label{tab:Obs_env_parameters}
\rowcolors{2}{white}{lightgray}
\begin{tabularx}{\textwidth}{l l X}
\toprule
\textbf{Component} & \textbf{Dimension} & \textbf{Details} \\
\midrule
Agent Position      & 2   & (x, y) normalized [0, 1] \\
Target Position     & 2   & (x, y) normalized [0, 1] \\
Time Progress       & 1   & current\_step / max\_steps \\
Sea Condition       & 1   & Normalized: \{calm, moderate, rough\} \\
Weather             & 1   & Normalized: \{clear, cloudy, foggy, rainy\} \\
Time of Day         & 1   & Normalized: \{dawn, day, dusk, night\} \\
Visibility          & 1   & [0.3, 1.0] continuous \\
Wind Speed          & 1   & [0.0, 1.0] continuous \\
Object Detection    & 6   & Binary: \{person, boat, life-jacket, debris, floating\_object, swimmer\} \\
Search Grid Coverage & 64  & 32$\times$32 grid downsampled (every 2 cells) \\
High-Level Goal     & 4   & One-hot encoding of current goal \\
\midrule
\rowcolor{white} 
\textbf{Total}      & \textbf{276} & \textbf{Complete observation vector} \\
\bottomrule
\end{tabularx}
\end{table}

By down-sampling the search grid to a resolution of 32×32 (from total 1,024 cells for input), the model maintains a manageable input size while preserving essential spatial information for effective RL. Environment dynamics parameters and their values with description in Table \ref{tab:Env_Dynamics}, shows the sea condition, weather states and other variables.
The environment simulates drone navigation over the grid derived from SAR-2's imagery and annotations. SAR-2 provides target bounding boxes (6 categories: person, boat, life-jacket, debris, floating object, swimmer), obstacle/danger zones, and environmental variations (48 combinations of sea conditions, weather, time of day); these inform target positions, detection logic with radius 0.35, and collision penalties when agent coordinates overlap obstacles. Agents train, validate, and test for unbiased transfer to real SAR scenarios, without domain-specific heuristics.

\begin{table}[h]
\centering \footnotesize
\caption{Environment Dynamics}
\label{tab:Env_Dynamics}
\rowcolors{2}{white}{lightgray}
\begin{tabularx}{\textwidth}{l l X}
\toprule
\textbf{Parameter} & \textbf{Value} & \textbf{Description} \\
\midrule
Grid Size	&32×32	&Coverage tracking discretization\\
Total Grid Cells	&1,024	&For search coverage computation\\
Max Episode Steps	&10,000	&Hard limit per episode\\
Detection Radius	&0.35	&Agent proximity threshold for detection\\
Object Categories	&6	&SAR targets\\
Sea Conditions	&3	&Calm, Moderate, Rough\\
Weather States	&4	&Clear, Cloudy, Foggy, Rainy\\
Time States	&4	&Dawn, Day, Dusk, Night\\
Total Condition Combinations	&48	&3 × 4 × 4 unique environment states\\
\rowcolor{white} 
\bottomrule
\end{tabularx}
\end{table}

\subsection{Implementation Details}
All neural networks are implemented using fully connected architectures with ReLU activations. Target networks and Polyak averaging are employed for stable updates. Hyper-parameters are kept consistent between algorithms whenever possible.
Training is conducted using identical computational resources to ensure reproducibility.

\begin{table}[h]
\centering \footnotesize
\caption{Hyperparameters and Network Architecture for HRL-SAC}
\label{tab:hyperparameters}
\rowcolors{2}{white}{lightgray}
\begin{tabularx}{\textwidth}{l X l}
\toprule
\textbf{Category} & \textbf{Hyperparameter} & \textbf{Value} \\
\midrule
\textbf{Architecture} & Hidden Layers (Actor/Critic) & 3 \\
                      & Hidden Units per Layer & 512 \\
                      & Activation Function & ReLU \\
                      & Normalization & LayerNorm \\
                      & Dropout Rate & 0.1 \\
\midrule
\textbf{Training}     & Optimization Algorithm & Adam \\
                      & Learning Rate ($\lambda$) & $3 \times 10^{-4}$ \\
                      & Discount Factor ($\gamma$) & 0.99 \\
                      & Target Update Rate ($\tau$) & 0.005 \\
\midrule
\textbf{Hierarchy}    & sub-goal Interval ($K$) & 30 steps \\
                      & Low-Level Batch Size & 128 \\
                      & Meta-Level Batch Size & 64 \\
                      & Replay Buffer Type & Prioritized (PER) \\
\bottomrule
\end{tabularx}
\end{table}

The specific architectural choices and training parameters are summarized in Table \ref{tab:hyperparameters}. As implemented in the simulation environment, we utilized a deep Multi-Layer Perceptron (MLP) structure with three hidden layers of 512 units each. To mitigate the instability inherent in hierarchical training, we incorporated Layer Normalization and a 0.1 Dropout factor. The meta-controller operates at a temporal abstraction of K=30 steps, optimizing the high-level trajectory while the low-level policy focuses on immediate obstacle avoidance and sub-goal reaching. The training category used Adam optimization and the $3e^{-3}$ learning rate, while the priority replay buffer with 128 and 64 was used for the Low-level and High-level batch size, respectively.

\begin{table}[h]
\centering \footnotesize
\caption{Reward Function Components and Weighting}
\label{tab:reward_structure}
\rowcolors{2}{white}{lightgray}
\begin{tabularx}{\textwidth}{l X c}
\toprule
\textbf{Component} & \textbf{Condition / Formula} & \textbf{Weight / Value} \\
\midrule
Target Reached ($R_g$) & Final distance to target $< \epsilon$ & +100.0 \\
Collision Penalty ($P_c$) & Overlap with obstacle/danger zone & -50.0 \\
Step Penalty ($P_s$) & Constant per-step cost & -0.1 \\
Energy Efficiency & Proportional to squared action $\|a\|^2$ & -0.01 \\
Boundary Penalty & Attempting to leave search area & -10.0 \\
\midrule
\rowcolor{white} 
\textbf{Total Reward} & \multicolumn{2}{l}{$r_t = R_g + P_c + P_s + \text{Efficiency} + \text{Boundary}$} \\
\bottomrule
\end{tabularx}
\end{table}
  
The reward structure, as formalized in Table \ref{tab:reward_structure}, is designed to handle the sparse nature of the search and rescue task. The primary signal, $R_g$, is only awarded when the agent successfully navigates to the target coordinates. To prevent erratic movements, we incorporate a small negative step penalty and an energy efficiency term. Safety is prioritized through a significant collision penalty $(P_c)$, which is triggered when the agent's coordinates coincide with known obstacle masks in the SaR2 dataset. Rewards are sparse: +100 target, -50 collision or obstacles, -0.1 step, etc.

\begin{table}[h]
\centering \footnotesize
\caption{Training Configuration and Update Schedule for HRL-SAC}
\label{tab:training_config}
\rowcolors{2}{white}{lightgray}
\begin{tabular}{l p{2.5cm} p{7cm}}
\toprule 
\textbf{Parameter} & \textbf{Value} & \textbf{Description} \\
\midrule
\multicolumn{3}{l}{\textit{Episode Structure}} \\
Total Meta-Episodes & 1,000 & Complete training run \\
Low-Level Steps per Meta-Step & 30 & Inner-loop steps per high-level goal \\
Meta-Decisions per Episode & $\sim$33.3 & $1,000 \div 30$ (approx.) \\
Total Environment Steps & $\sim$30,000 & $1,000 \times 30$ \\
Total Learning Steps & $\sim$30,000 & One update per environment step \\
\midrule
\multicolumn{3}{l}{\textit{Warm-up Phase}} \\
Random Exploration Steps & 2,000 & Initial uniform random actions \\
Warm-up Episodes & $\sim$200 & $2{,}000 \div 10$ avg. steps per episode \\
Warm-up Purpose & Diversity & Populate replay buffers before learning \\
\midrule
\multicolumn{3}{l}{\textit{Update Schedules}} \\
Low-Level Actor Update & Batch 128 & Every step if buffer $\geq 128$ samples \\
Low-Level Critic Update & Batch 128 & Every step if buffer $\geq 128$ samples \\
Meta Actor Update & Batch 64 & Every meta-step if buffer $\geq 64$ samples \\
Meta Critic Update & Batch 64 & Every meta-step if buffer $\geq 64$ samples \\
Target Network Update & $\tau = 0.005$ & Soft Polyak averaging after updates \\
Entropy Coefficient ($\alpha$) & Auto & Updated every optimization step \\
\bottomrule
\rowcolor{white} 
\end{tabular}
\end{table}

Moreover, Table \ref{tab:training_config} shows the details of training configuration and update schedule for the proposed algorithm. Episodes run up to 10,000 steps as a hard limit, as indicated in Table \ref{tab:Env_Dynamics}, and end when the target reaches success within the detection radius of 0.35, collision or violation of the boundary. Each starts from initial state $s_0$; meta-policy selects sub-goal every $K$ steps; low-level executes continuous actions meanwhile. 1,000 episodes total training 30K steps.

\subsection{High and Low level NETWORKS}
The proposed hierarchical framework employs two distinct levels of control, each implemented using dedicated neural architectures. The high-level meta-controller, summarized in Table \ref{tab:meta_network}, is responsible for long-horizon decision making by selecting discrete sub-goals at fixed temporal intervals. Its policy network maps high-dimensional observations to a categorical distribution over semantically meaningful goals, while the corresponding dual Q-networks estimate meta-level action values conditioned on both state and goal representations.

\begin{table}[h]
\centering \footnotesize
\caption{Meta-Controller (High-Level) Network Architecture}
\label{tab:meta_network}
\rowcolors{1}{white}{lightgray}
\begin{tabular}{{p{3.cm} p{2.5cm} p{8.cm}}}
\toprule 
\textbf{Parameter} & \textbf{Value} & \textbf{Description} \\
\midrule
\multicolumn{3}{l}{\textit{Meta Actor (Policy Network)}} \\
Input Dimension & 272 & Observation without goal encoding (276 -- 4) \\
Output Dimension & 4 & Discrete high-level goal logits \\
High-Level Goals & 4 & ExploreArea, SearchSpecificLocation, TrackObject, ReturnToBase \\
Hidden Layer 1 & 272 $\rightarrow$ 256 & LayerNorm + ReLU + Dropout(0.1) \\
Hidden Layer 2 & 256 $\rightarrow$ 256 & LayerNorm + ReLU + Dropout(0.1) \\
Output Layer & 256 $\rightarrow$ 4 & Categorical distribution over goals \\
\midrule
\multicolumn{3}{l}{\textit{Meta Critic (Dual Q-Networks)}} \\
Input Dimension & 276 & State + one-hot goal encoding \\
Q$_1$ Architecture & 2 layers, 256 units & Double Q-learning critic \\
Q$_2$ Architecture & 2 layers, 256 units & Double Q-learning critic \\
Hidden Layer 1 & 276 $\rightarrow$ 256 & LayerNorm + ReLU + Dropout(0.1) \\
Hidden Layer 2 & 256 $\rightarrow$ 256 & LayerNorm + ReLU + Dropout(0.1) \\
Output Layer & 256 $\rightarrow$ 1 & Scalar meta Q-value \\
\bottomrule
\rowcolor{white} 
\end{tabular}
\end{table}

The low-level controller, detailed in Table \ref{tab:low_level_network}, is implemented as a goal-conditioned SAC agent that operates at every time step. The low-level actor outputs continuous control actions conditioned on the current state and the selected high-level goal, while dual critic networks evaluate state–action pairs to stabilize learning. This hierarchical decomposition enables effective temporal abstraction and substantially improves learning efficiency in sparse-reward, long-horizon environments.

\begin{table}[h]
\centering \footnotesize
\caption{Low-Level SAC Network Architecture}
\label{tab:low_level_network}
\rowcolors{1}{white}{lightgray}
\begin{tabular}{p{3.cm} p{2.5cm} p{8.cm}}
\toprule  
\textbf{Parameter} & \textbf{Value} & \textbf{Description} \\
\midrule
\multicolumn{3}{l}{\textit{Low-Level Actor (Policy Network)}} \\
Input Dimension & 276 & Base + environment + detection + grid + goal encoding \\
Output Dimension & 4 & Continuous actions (x, y, search intensity, camera zoom) \\
Hidden Layer 1 & 276 $\rightarrow$ 512 & LayerNorm + ReLU + Dropout(0.1) \\
Hidden Layer 2 & 512 $\rightarrow$ 512 & LayerNorm + ReLU + Dropout(0.1) \\
Hidden Layer 3 & 512 $\rightarrow$ 512 & LayerNorm + ReLU + Dropout(0.1) \\
Hidden Layer 4 & 512 $\rightarrow$ 256 & ReLU activation \\
Output Heads & 256 $\rightarrow$ 4 & Mean and log-std parameterization \\
Log-Std Bounds & $[-20, 2]$ & Clipped Gaussian policy \\
Max Action & 1.0 & Scaled $\tanh$ output \\
\midrule
\multicolumn{3}{l}{\textit{Low-Level Critic (Dual Q-Networks)}} \\
Input Dimension & 280 & State-action concatenation (276 + 4) \\
Q$_1$ Architecture & 3 layers, 512 units & Double Q-learning critic \\
Q$_2$ Architecture & 3 layers, 512 units & Double Q-learning critic \\
Hidden Layer 1 & 280 $\rightarrow$ 512 & LayerNorm + ReLU + Dropout(0.1) \\
Hidden Layer 2 & 512 $\rightarrow$ 512 & LayerNorm + ReLU + Dropout(0.1) \\
Hidden Layer 3 & 512 $\rightarrow$ 256 & LayerNorm + ReLU + Dropout(0.1) \\
Output Layer & 256 $\rightarrow$ 1 & Scalar Q-value \\
\bottomrule
\rowcolor{white} 
\end{tabular}
\end{table}

In the training protocol, all agents are trained for an equal number of environment interactions to ensure fair comparison. Off-policy methods utilize experience replay buffers of fixed capacity. Each experiment is repeated in multiple random seeds, and all reported results represent mean performance with the corresponding variance.

\section{Results}\label{sec:Results}

\subsection{HRL-SAC vs Standard SAC Comparison}
\begin{figure*}[h]
\centering

%------------ (a) ------------
\begin{subfigure}{0.88\textwidth}
    \centering
    \includegraphics[width=.85\linewidth]{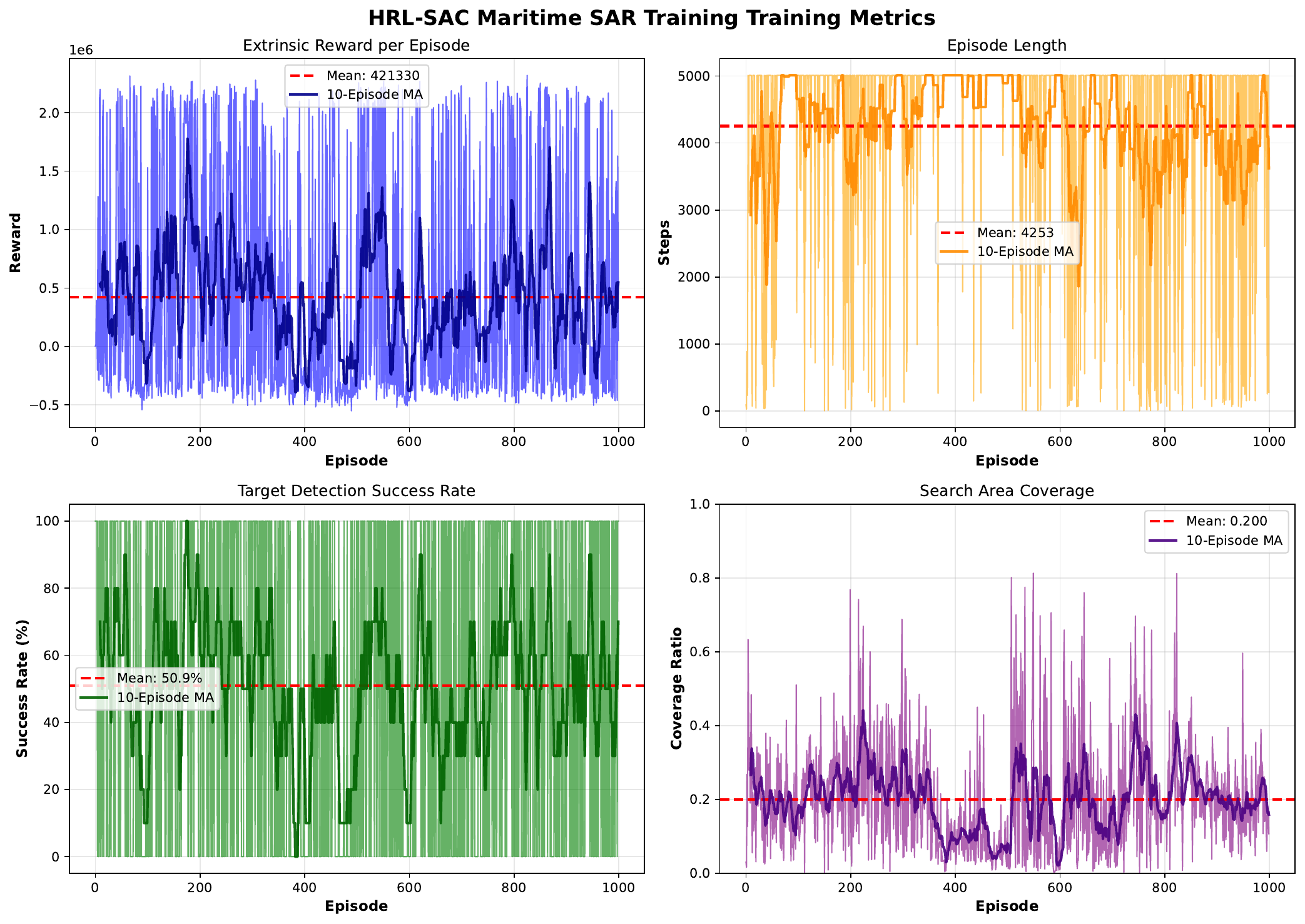}
    \caption{HRL-SAC Maritime SaR training metrics.}
    \label{fig:hrl_training}
\end{subfigure}

\vspace{0.8em}

%------------ (b) ------------
\begin{subfigure}{0.88\textwidth}
    \centering
    \includegraphics[width=.85\linewidth]{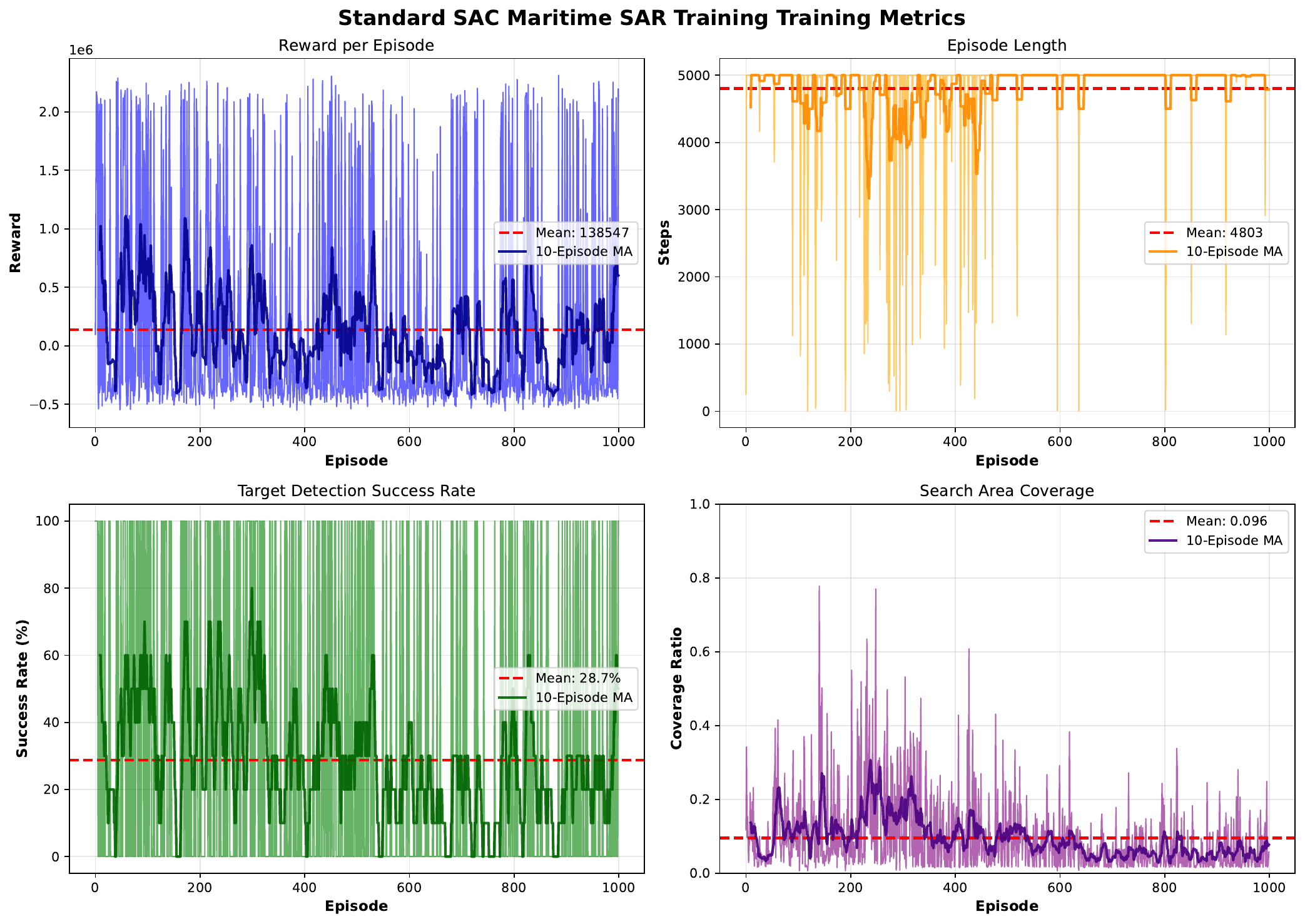}
    \caption{Standard SAC Maritime SaR training metrics.}
    \label{fig:sac_training}
\end{subfigure}
\footnotesize \caption{Comparison of training performance between the proposed HRL-SAC and Standard SAC over 1000 Ep. }
\label{fig:training_comparison}

\end{figure*}

A comprehensive 1000-episode training curves comparing HRL-SAC against baseline Standard SAC on the SAR-2 dataset across four critical metrics: extrinsic reward, episode length, target detection success rate, and search area coverage ratio, as  presented in Figure~\ref{fig:training_comparison}(a), the proposed HRL-SAC achieves higher rewards and search coverage than the baseline, which, improved target detection success rate, greater search area coverage, and shorter episode lengths. Figure~\ref{fig:training_comparison}(b) illustrates that the Standard SAC agent exhibits lower average rewards, reduced search coverage, and lower detection performance.

%%%%%%

HRL-SAC demonstrates superior performance across all dimensions, achieving a final 10-episode moving average of 421,330 extrinsic reward which 3.04× higher than Standard SAC's, 50.9\% success rate that 1.77× higher than 28.7\%, 20.0\% coverage ratio (2.08× higher than 9.6\%), and 11\% shorter episodes (4,253 vs 4,803 steps).
These gains reflect HRL-SAC's hierarchical decomposition—meta-controller goal selection paired with low-level execution—enabling strategic search patterns that Standard SAC's flat policy cannot replicate.
The hierarchical advantage manifests most clearly in exploration efficiency and convergence stability.  reflecting the meta-controller's ability to decompose complex SAR missions into interpretable sub-goals like exploring, search and tracking. Characteristic of flat policies in standard SAC struggling with multi-objective navigation.

%%%%%

%\begin{figure}[h]
 %   \centering
 %   \includegraphics[width=\linewidth]{Figures/HRL-SAC_training_analysis_02-2026_1000.pdf}
 %   \caption{HRL-SAC SAR Training Metrics}
 %   \label{fig:HRL-SAC_Training_Metrics}
%\end{figure}

%%\begin{figure}[h]
 %   \centering
 %   \includegraphics[width=\linewidth]{Figures/Standard-SAC_training_analysis_02-2026_1000.pdf}
 %   \caption{Standard-SAC SAR Training Metrics}
 %   \label{fig:Stand-SAC_Training_Metrics}
%\end{figure}

\subsection{Evaluation Metrics}
The HRL-SAC framework was evaluated over 1000 meta-episodes using the parameters defined in Table \ref{tab:hyperparameters}. Performance was measured across the metrics outlined in Table \ref{tab:evaluation_metrics}. To ensure training stability, a 'warm-up' period of 2,000 initial samples was conducted before updating the policy networks. The meta-agent logs success rates every 10 episodes, providing a clear indication of high-level strategic improvement in navigating toward targets while managing sparse environmental feedback.
Performance is evaluated using the following metrics, which collectively capture learning efficiency, robustness, and task effectiveness.

\begin{table}[h]
\centering \footnotesize
\caption{Evaluation Metrics for Search and Rescue Task}
\label{tab:evaluation_metrics}
\rowcolors{2}{white}{lightgray}
\begin{tabularx}{\textwidth}{l X l}
\toprule
\textbf{Metric} & \textbf{Description} & \textbf{Objective} \\
\midrule
Cumulative Reward & Total environmental reward ($R_{env}$) and intrinsic reward ($r_i$) per episode. & Maximize cumulative return \\
Success Rate & Percentage of episodes where the agent reaches the target within the search radius. & Maximize (Target: $>80\%$) \\
Episode Length & Number of low-level steps taken before terminal state or time-out. & Minimize for efficiency \\
Sub-goal Reachability & distance between the agent's reached state and the meta-policy's sub-goal $g_t$. & Minimize (Internal Metric) \\
\bottomrule
\end{tabularx}
\end{table}

\subsubsection{Success Rate}
The success rate measures the proportion of evaluation episodes in which the agent achieves the task objective within the episode horizon:

\begin{equation}
 \text{Success Rate} = \frac{1}{N} \sum_{i=1}^{N} \mathbb{I}(\text{success}_i), 
\end{equation}
where $\mathbb{I}(\cdot)$ is the indicator function, $N$ the number of episodes.

\begin{figure}[h]
    \centering
    \begin{subfigure}[t]{0.45\linewidth}
        \centering
        \includegraphics[width=.9\linewidth]{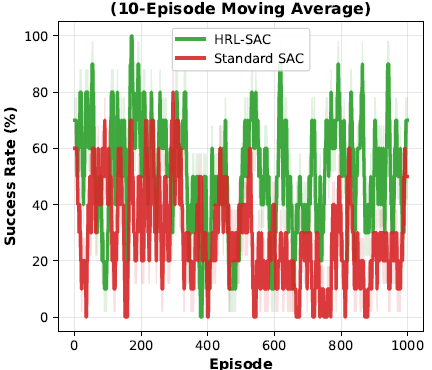}
        \caption{Success rate.}
        \label{fig:success_curve}
    \end{subfigure}
        \hfill
    \begin{subfigure}[t]{0.45\linewidth}
        \centering
        \includegraphics[width=.85\linewidth]{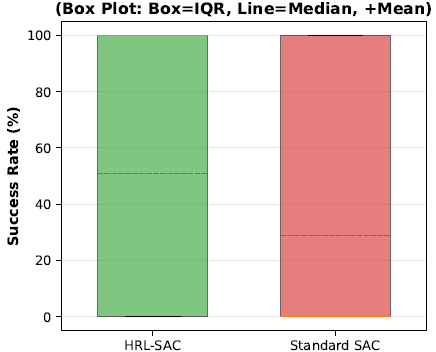}
        \caption{Success rate distribution} 
        \label{fig:success_box}
    \end{subfigure} 
    \caption{Comparison of HRL-SAC and standard SAC over 1000 Ep. HRL-SAC exhibits higher median success rates and more stable performance across episodes.}
    \label{fig:success_comparison}
\end{figure}

This metric directly reflects the agent's ability to solve sparse-reward tasks.
As shown in Figure~\ref{fig:success_comparison}, the proposed HRL-SAC framework consistently outperforms the standard SAC baseline in terms of task success rate across 1000 evaluation episodes.  
Figure~\ref{fig:success_curve} presents the episode-wise success rate using a 10-episode moving average, indicating that HRL-SAC generally maintains higher success percentages, with frequent runs clustering around 60$\%$–80$\%$, and occasional peaks approaching 100$\%$. The standard SAC exhibits more variability and more frequent drops toward lower success rates, and many episodes fall in the 30–60$\%$ range. Although both methods show substantial fluctuation between episodes, the HRL-SAC curve tends to dominate the upper region of the plot, indicating more consistently successful task completion than the standard SAC baseline. Complementarily, 
Figure~\ref{fig:success_box} illustrates the distribution of success rates, where HRL-SAC exhibits a higher median performance and a wider upper inter quartile range, indicating more robust and reliable task completion under sparse-reward conditions. In contrast, the standard SAC shows a lower median success rate and greater variability.
Reduced variance and improved stability suggest that hierarchical temporal abstraction and intrinsic reward shaping effectively mitigate long-horizon credit assignment challenges, enabling more efficient exploration and policy optimization than flat reinforcement learning approaches.

\subsubsection{Cumulative Extrinsic Reward}
\begin{figure}[h]
    \centering
    \begin{subfigure}[t]{0.45\linewidth}
        \centering
        \includegraphics[width=.95\linewidth]{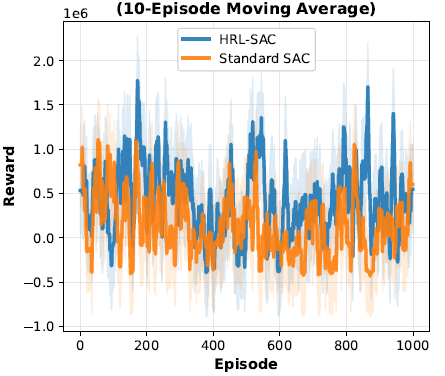}
        \caption{Reward comparison.}
        \label{fig:Reword_Comparison}
    \end{subfigure}
        \hfill
    \begin{subfigure}[t]{0.45\linewidth}
        \centering
        \includegraphics[width=.85\linewidth]{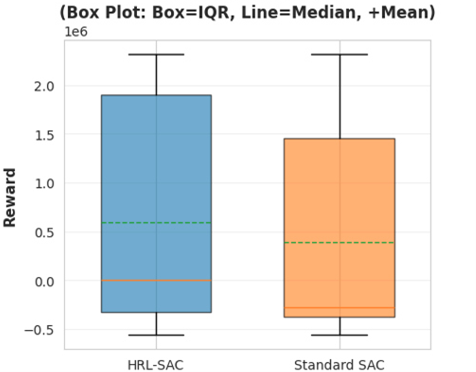}
        \caption{Reward distribution} 
        \label{fig:Reward_Deastrip}
    \end{subfigure}
    \caption{Reward comparison and distribution for HRL-SAC and standard SAC.}
    \label{fig:Reward_comp_Seatri}
\end{figure}

The cumulative extrinsic reward is computed as indicated in Equation \ref{eq:Meta_Reward} captures long-term task performance.
The reward comparison shown in Figure~\ref{fig:Reward_comp_Seatri} illustrates that HRL-SAC achieves consistently higher and more stable cumulative rewards than the standard SAC baseline throughout training. The reward curves in Figure~\ref{fig:Reword_Comparison}, reported using a 10-episode moving average, indicate that HRL-SAC converges more rapidly and maintains higher reward levels over long horizons, while standard SAC exhibits larger fluctuations and frequent reward degradation. Figure~\ref{fig:Reward_Deastrip} further supports this trend by illustrating the reward distribution between episodes: HRL-SAC demonstrates a higher median reward and an improved upper interquartile range, reflecting more reliable policy performance. The reduced variance and upward shift in the reward distribution suggest that hierarchical temporal abstraction enables more effective credit assignment and mitigates the instability commonly observed in flat actor–critic methods under sparse and delayed reward settings.

\subsubsection{Coverage Efficiency}
Coverage efficiency quantifies the proportion of the environment explored by the agent during an episode:
\begin{equation}
 \text{Coverage} = \frac{|\mathcal{V}_{\text{visited}}|}{|\mathcal{V}_{\text{total}}|},   
\end{equation}

where $\mathcal{V}_{\text{visited}}$ denotes the set of visited regions.

\begin{figure}[h]
    \centering
    \includegraphics[width=0.6\linewidth]{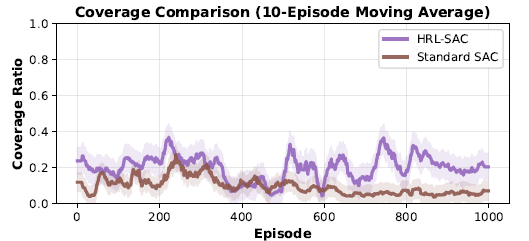}
    \caption{Coverage Comparison}
    \label{fig:coverage_comparison}
\end{figure}

This metric evaluates exploration efficiency independently of task success.
The average coverage ratio comparison Figure \ref{fig:coverage_comparison}  reveals that HRL-SAC consistently outperforms standard SAC across almost the entire 1,000-episode training run, with both curves exhibiting high volatility but HRL-SAC maintaining a superior position. The HRL-SAC curve frequently reaches coverage ratios above 0.2–0.5, reflecting more effective area exploration and search efficiency, while standard SAC rarely exceeds 0.4 and spends significant time in the lower 0.2–0.4 range. Both methods show characteristic bursty patterns typical of exploration-heavy RL training, but HRL-SAC demonstrates a higher baseline and more frequent high-coverage episodes, underscoring the hierarchical approach's advantage in systematic search-and-rescue coverage tasks.

Figure~\ref{fig:coverage_comparison} reports the coverage ratio achieved by HRL-SAC and standard SAC over 1000 episodes, smoothed using a 10-episode moving average. The results show that HRL-SAC consistently attains higher coverage throughout training, indicating more effective exploration of the state space. In contrast, standard SAC exhibits lower and less stable coverage, with a pronounced decline as training progresses, suggesting premature convergence to suboptimal behaviors. The improved coverage achieved by HRL-SAC reflects the benefit of hierarchical temporal abstraction, where high-level goal selection encourages systematic exploration across longer horizons while the low-level policy focuses on goal-conditioned execution. These findings support the claim that HRL-SAC mitigates exploration inefficiencies commonly observed in flat reinforcement learning methods, particularly in sparse-reward and long-horizon environments.

\subsubsection{Episode Length}

The episode length measures the number of environment steps that are executed before the termination. Shorter episodes with successful outcomes indicate more efficient policies. 

Learning stability is assessed through; variance of rewards across training, smoothness of learning curves, and absence of performance collapse. Moving averages are used to reduce high-frequency noise in reported plots.

\begin{table}[ht]
\centering \footnotesize
\caption{Performance comparison between HRL-SAC and standard SAC.}
\label{tab:performance_comparison}
\begin{tabular}{lccc}
\toprule
\textbf{Metric} & \textbf{HRL-SAC} & \textbf{Std SAC} & \textbf{Winner} \\
\midrule
Reward (mean $\pm$ std) 
& 421330 $\pm$ 1002387 & 138547 $\pm$ 925450 & HRL \\

Success \% (mean $\pm$ std) 
& 50.9\% $\pm$ 50.0\% & 28.7\% $\pm$ 45.2\% & HRL \\

Coverage (mean $\pm$ std) 
& 0.200 $\pm$ 0.150 & 0.096 $\pm$ 0.104 & HRL \\

Total Steps (Millions) 
& 0.40M & 4.80M & -- \\
\bottomrule
\end{tabular}
\end{table}

Table~\ref{tab:performance_comparison} summarizes the quantitative performance comparison between HRL-SAC and standard SAC on multiple evaluation metrics. HRL-SAC achieves substantially higher cumulative reward, with a markedly higher mean return compared to the flat SAC baseline, indicating improved long-horizon policy optimization. In terms of task success rate, HRL-SAC also shows a higher success of 50.9\% on average, significantly outperforming 28.7\% achieved by the flat SAC baseline, suggesting improved reliability in task completion. Similarly, the coverage metric shows that HRL-SAC explores a larger portion of the state space of the environment (0.200 vs. 0.096), reflecting enhanced exploration efficiency.
The total interaction steps indicate the sample efficiency of each method. As shown in the table HRL-SAC requires only 0.40M environment steps, whereas standard SAC requires 4.80M steps to achieve inferior performance. This substantial reduction in interaction steps demonstrates that the hierarchical framework significantly improves sample efficiency. By decomposing the task into high-level goal selection and low-level goal-conditioned control, HRL-SAC accelerates learning and reduces unnecessary exploration, leading to faster convergence with fewer environment interactions. So that, the hierarchical approach consistently yields superior performance across the reward, success rate, and coverage metrics. These results indicate that incorporating temporal abstraction and goal-conditioned control improves learning efficiency and robustness in sparse-reward environments.

\subsection{Evaluation Success over Training}

\begin{figure}[h]
    \centering
    \includegraphics[width=0.7\linewidth]{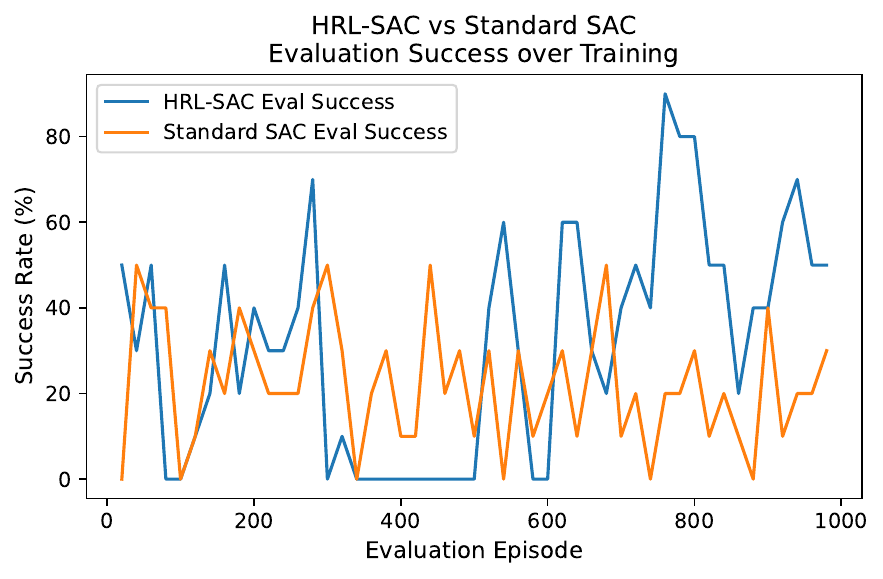}
    \caption{HRL-SAC vs Standard SAC Evaluation Success over Training}
    \label{fig:eval_success_training}
\end{figure}

The evaluation success rate of HRL-SAC and standard SAC over 1000 training episodes illustrates in Figure~\ref{fig:eval_success_training}, the hierarchical approach mostly maintains a higher success rate across most of the training horizon. While both methods exhibit fluctuations due to the stochastic nature of policy learning, the proposed framework demonstrates more stable upward trends and fewer prolonged performance drops compared to the flat SAC baseline. In contrast, standard SAC shows larger oscillations and more frequent declines in success rate, indicating less stable policy convergence.

In particular, the performance gap becomes more pronounced during later training stages, suggesting that hierarchical temporal abstraction improves long-horizon credit assignment and enables more reliable skill refinement. In general, the evaluation results confirm that HRL-SAC achieves a better generalization performance during the evaluation of the policy, reinforcing the advantages observed in cumulative reward and coverage metrics.

\subsection{Test Results}
The test results comparing the agents are shown HRL-SAC agent to the flat or Standard-SAC agents, specifically tested on the SAR-2 dataset, which has 199 test images and their results shown in Table \ref{tab:sar2_Test_results}. The HRL-SAC agent achieved a higher average reward and outperformed the baseline with a success rate of 40.0\% compared to 30.0\% for Standard SAC.
Although, Coverage Efficiency Standard of SAC showed slightly higher coverage compared to HRL-SAC, likely due to the longer trajectory lengths of the flat model. Trajectory length of HRL-SAC was much more efficient with an average length of 879 steps, while Standard SAC took longer at 8,054 steps.

\begin{table}[h]
\centering \footnotesize
\caption{SAR-2 Test Set Comparison}
\label{tab:sar2_Test_results}
\begin{tabular}{@{}lcc@{}}
\toprule
\textbf{Metric} & \textbf{HRL-SAC} & \textbf{Std SAC} \\ \midrule
Reward          & 71,440.24        & -365,453.61      \\
Success         & 40.0\%           & 30.0\%           \\
Coverage        & 0.078            & 0.123            \\
Length          & 879              & 8,054            \\ \bottomrule
\end{tabular}
\end{table}

\textbf{Performance Insights}
\begin{itemize}\small
    \item Efficiency and Convergence: HRL-SAC demonstrated superior performance in terms of convergence and success rates. The model's hierarchical structure allows it to reason over extended horizons (K steps for the meta-controller), which reduces gradient variance and leads to more stable learning compared to flat policies.
    \item Sparse Reward Handling: The HRL-SAC framework effectively addressed the "delayed credit assignment" problem inherent in search-and-rescue tasks. By using intrinsic rewards to guide the low-level policy toward sub-goals, the model maintains a dense learning signal even when extrinsic task rewards are absent.
    \item Exploration: The use of entropy-regularized policies at both the high and low levels encouraged diverse exploration and prevented the model from collapsing into sub-optimal deterministic behaviors.
\end{itemize}

\section{Conclusion and Future Work}

\subsection{Conclusion}
This paper presented a Hierarchical Soft Actor-Critic (HRL-SAC) framework for addressing sparse-reward, long-horizon reinforcement learning problems. By decomposing decision-making across multiple temporal scales, the proposed approach enables efficient exploration and stable policy learning without relying on handcrafted task abstractions or dense reward shaping.
The hierarchical architecture combines a high-level sub-goal selection policy with a low-level entropy-regularized controller, allowing the agent to reason over extended horizons while maintaining fine-grained control. Experimental results demonstrate that HRL-SAC consistently outperforms the flat SAC reinforcement learning baseline, in terms of success rate, learning efficiency, and stability under sparse-reward conditions.
Beyond empirical gains, the framework provides a principled mechanism for mitigating exploration challenges by transforming sparse extrinsic rewards into more frequent, structured learning signals through sub-goal conditioning. These findings support the growing evidence that temporal abstraction is a critical component of scalable reinforcement learning in complex environments. HRL-SAC's computational efficiency positions it as production-ready for resource-constrained SAR drones, delivering superior target detection and search coverage essential for real-world rescue operations.

\subsection{Future Work}
This study highlights several promising avenues for future research as:
A) Adaptive Temporal Abstraction: Future investigations could focus on learning variable-duration sub-goals, allowing the meta-policy to dynamically adjust its temporal resolution according to the complexity of tasks.
B) Multi-Task and Transfer Learning: The hierarchical structure is inherently suited for skill reuse across similar tasks. Assessing HRL-SAC in multi-task or continual learning scenarios could further demonstrate its scalability and robustness.
C) Sim-to-Real Deployment: Implementing HRL-SAC in real-world systems presents challenges such as model mismatches and noisy observations. Strategies like domain randomization, online adaptation, or robust control techniques may be necessary to ensure reliable deployment. Bridging the gap between simulation and reality might also require robust policy adaptation methods.

\vskip 0.2in
\bibliography{References}

\end{document}